\documentclass[10pt,twocolumn,letterpaper]{article}

 \usepackage{iccv}              

\usepackage{gensymb}

\definecolor{iccvblue}{rgb}{0.21,0.49,0.74}
\usepackage[pagebackref,breaklinks,colorlinks,allcolors=iccvblue]{hyperref}

\def\paperID{8} 
\def\confName{ICCV}
\def\confYear{2025}

\title{Ego-Exo 3D Hand Tracking in the Wild with a Mobile Multi-Camera Rig}

\author{
Patrick Rim\textsuperscript{1,2}\quad
Kun He\textsuperscript{1}\quad
Kevin Harris\textsuperscript{1}\quad
Braden Copple\textsuperscript{1}\quad
Shangchen Han\textsuperscript{1}\quad
Sizhe An\textsuperscript{1}\\
Ivan Shugurov\textsuperscript{1}\quad
Tomas Hodan\textsuperscript{1}\quad
He Wen\textsuperscript{1}\quad
Xu Xie\textsuperscript{1}\\[0.2cm]
\textsuperscript{1}Meta Reality Labs\quad
\textsuperscript{2}Yale University\\
}

\begin{document}
\maketitle

\begin{abstract}
Accurate 3D tracking of hands and their interactions with the world in unconstrained settings remains a significant challenge for egocentric computer vision. 
With few exceptions, existing datasets are predominantly captured in controlled lab setups, 
limiting environmental diversity and model generalization. 
To address this, we introduce a novel marker-less multi-camera system designed to capture precise 3D hands and objects, which allows for nearly unconstrained mobility in genuinely \textbf{in-the-wild} conditions. 
We combine a lightweight, back-mounted capture rig with eight exocentric cameras, 
and a user-worn Meta Quest 3 headset, which contributes two egocentric views. 
We design an ego-exo tracking pipeline to generate accurate 3D hand pose ground truth from this system, and rigorously evaluate its quality. 
By collecting an annotated dataset featuring synchronized multi-view images and precise 3D hand poses, 
we demonstrate the capability of our approach to significantly reduce the trade-off between environmental realism and 3D annotation accuracy. 
\end{abstract}

\section{Introduction}
\label{sec:intro}

Accurate 3D tracking of human hands and their interactions with objects in everyday environments is a foundational capability for 
egocentric perception systems. 
In augmented and virtual reality (AR/VR), hand-object perception supports natural and expressive user interactions without reliance on external input devices 
or constrained input spaces. 
And in robotics and teleoperation, observing how humans manipulate tools and objects in diverse settings is critical for learning transferable 
manipulation skills. 

Although recent advances in computer vision 
have significantly improved hand-object perception models~\cite{han2020megatrack, han2022umetrack, pavlakos2024reconstructing, zhang2016learning, yang2024mlphand}, they may still struggle when deployed outside of mostly controlled and indoor environments. 
Thus, there is a pressing need for data capture systems that can produce authentic, unconstrained data that reflects hand-object interactions in the wild. 
However, building such systems remains highly challenging due to a fundamental trade-off.
On the one hand, the need for authentic hand/object appearance, environmental diversity, and user mobility limits the application of external sensors, e.g., cameras, IMUs, ToF and motion capture systems~\cite{kim2009multi, sun20213drimr, xia2023quadric, alexiadis2016integrated}. On the other hand, the availability and quality of 3D ground truth are ultimately bounded by sensor fidelity and coverage.

\begin{figure*}[t]
  \centering
  \includegraphics[width=0.9\linewidth]{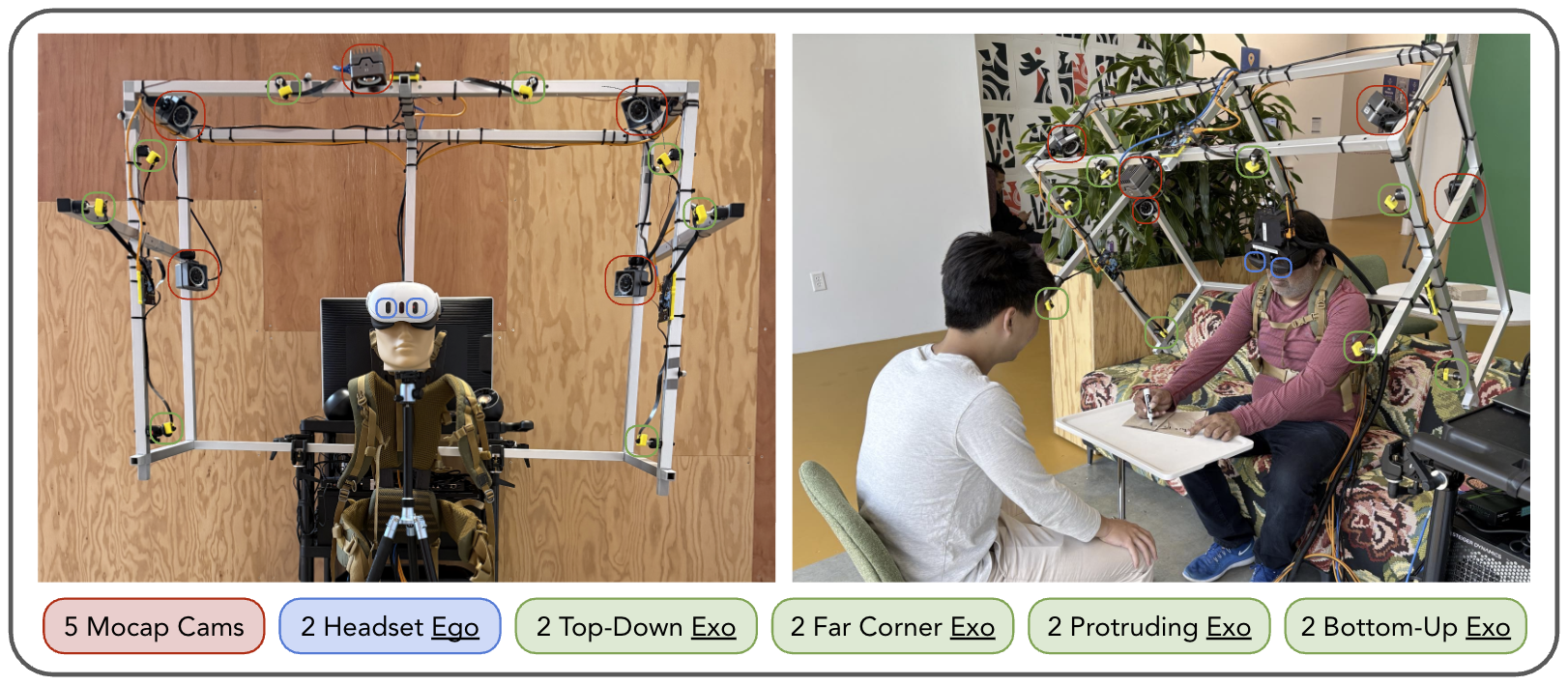}
  \vspace{-3mm}
  \caption{\textbf{Our mobile multi-camera capture rig.} Left: annotated hardware layout showing five OptiTrack motion capture cameras (red), two egocentric headset cameras (blue), and eight exocentric fisheye cameras mounted in a half-dome configuration (green) including top-down, far-corner, protruding, and bottom-up placements. Right: the rig in use during an in-the-wild capture session, demonstrating its lightweight, wearable design that allows natural interaction while maintaining synchronized multi-view coverage.}
  \label{fig:rig}
  \vspace{-3mm}
\end{figure*}

Existing large-scale hand-object interaction datasets often choose from the two extremes in this trade-off.
Most of such datasets, such as ARCTIC \cite{fan2023arctic} and HOT3D \cite{banerjee2025hot3d}, are captured under controlled conditions, typically within indoor laboratory spaces equipped with marker-based motion capture systems or stationary multi-camera arrays.
These setups can offer high 3D accuracy, but inherently limit the environmental diversity of collected data. 
At the other extreme, datasets with significant environmental variations, such as Ego-Exo 4D \cite{grauman2024ego} and Nymeria \cite{ma2024nymeria}, are limited in their sensor setup, and therefore lack dense and accurate 3D ground truth for hands and objects.


In this work, we take inspiration from the plethora of multi-camera systems in the literature, but design ours in a way such that we break the existing trade-off between realism and ground truth sensing. 
Our novel multi-camera rig is capable of acquiring accurate 3D ground truth, but also maintains high mobility such that we can capture authentic egocentric interactions in diverse environments, including outdoors. 
The setup comprises a lightweight backpack-style rig equipped with eight synchronized fisheye monochrome cameras that provide exocentric views, as well as 
a Meta Quest 3 headset worn by the user, providing two egocentric views from the front-facing fisheye cameras.
Currently, 
by applying a multi-stage ego-exo hand tracking pipeline utilizing all ten camera views, 
we obtain accurate 3D hand pose annotations in challenging environments that are mostly unseen in existing datasets. 

In this extended abstract, we introduce the core system, 
and describe the initial dataset that we collected and annotated.
Quantitatively, we evaluate the ground truth quality of our system, and show that the data we capture poses a higher level of challenge for egocentric hand pose estimation compared to other related datasets.

\section{Related Work}
\label{sec:relatedwork}

In this section, we review the most recent efforts in large-scale hand-object datasets, emphasizing their data collection setups and annotation methodologies.

HOI4D~\cite{liu2022hoi4d} records RGB-D sequences using a Kinect v2 and an Intel RealSense camera mounted on a bicycle helmet worn by participants in indoor environments. Ground truth labels for hands and objects are obtained using manual annotation and frame-wise propagation. Despite its pioneering scale, relying solely on a single egocentric view 
severely constrains the fidelity of ground truth, as it is challenging to resolve occlusions and scale ambiguity.

Marker-based motion capture (MoCap) systems have been used extensively for ground truth capture purposes.
ARCTIC~\cite{fan2023arctic} uses 
an off-the-shelf solution with 54 infrared (IR) MoCap cameras to track hands and objects by attaching IR-reflective markers to them.
However, the stationary dome-style setup limits environmental diversity, and the presence of markers detracts from visual realism and impedes natural interactions.
OakInk2~\cite{zhan2024oakink2} employs a hybrid capture solution with 12 MoCap cameras and four RGB cameras (three exocentric, one egocentric). Similar to ARCTIC, both the hands and the objects are embedded with markers so as to generate 3D pose ground truth.
TACO~\cite{liu2024taco} uses a setup consisting of 12 static exocentric RGB cameras, one egocentric GoPro camera, and a MoCap system. 
While object poses are still marker-based, hand pose annotations are derived from a marker-less pipeline: YOLOv3-based detection~\cite{redmon2018yolov3}, MMPose~\cite{sengupta2020mm} for 2D keypoints, and then 3D localization via triangulation. 
A recent dataset, HOT3D~\cite{banerjee2025hot3d}, works towards realism by utilizing the Meta Quest 3 headset and Aria research glasses \cite{engel2023projectarianewtool}.
However, the hands and objects are still tracked with a static MoCap system and rely on markers.

On the other hand, marker-less solutions typically rely on a synchronized multi-camera setup.
Assembly101~\cite{sener2022assembly101} and AssemblyHands~\cite{ohkawa2023assemblyhands} use 8 RGB cameras synchronized with a custom-built headset to capture hand-object interactions and annotate 3D hand poses, but lack ground-truth object pose. 
More recently, GigaHands~\cite{fu2025gigahands} 
utilizes an array of 51 RGB cameras to build a 2D-to-3D pipeline for hand and object pose annotation, involving YOLOv8~\cite{10533619} for detection, HaMeR~\cite{pavlakos2024reconstructing} for 2D keypoints, and triangulation followed by MANO fitting~\cite{romero2022embodied}. The dataset notably offers a large number of frames compared to existing datasets,
but the stationary and restricted capture volume still results in 
limited environmental and contextual diversity.
Furthermore, GigaHands does not contain any egocentric views. 

A notable large-scale dataset with real-world settings is Ego-Exo4D~\cite{grauman2024ego}, featuring one egocentric view and four exocentric views. However, since its primary focus is on capturing full-body actions from a distance greater than that typically used for observing hands, hand annotations are available only in a sparse subset of frames, obtained mostly using manual annotation. Moreover, Ego-Exo4D does not provide 3D annotations for object pose.
Nymeria~\cite{ma2024nymeria}, a similar real-world dataset captured with Aria research glasses and wearable MoCap suits, also primarily focuses on full-body motion instead of hand-object interactions.

Across all these existing datasets, an evident trade-off emerges: employing extensive multi-camera motion capture setups (marker-based or marker-less) can yield high-quality annotations, but sacrifices environmental diversity and visual realism. 

\begin{figure*}[t]
  \centering
  \includegraphics[width=1.0\linewidth]{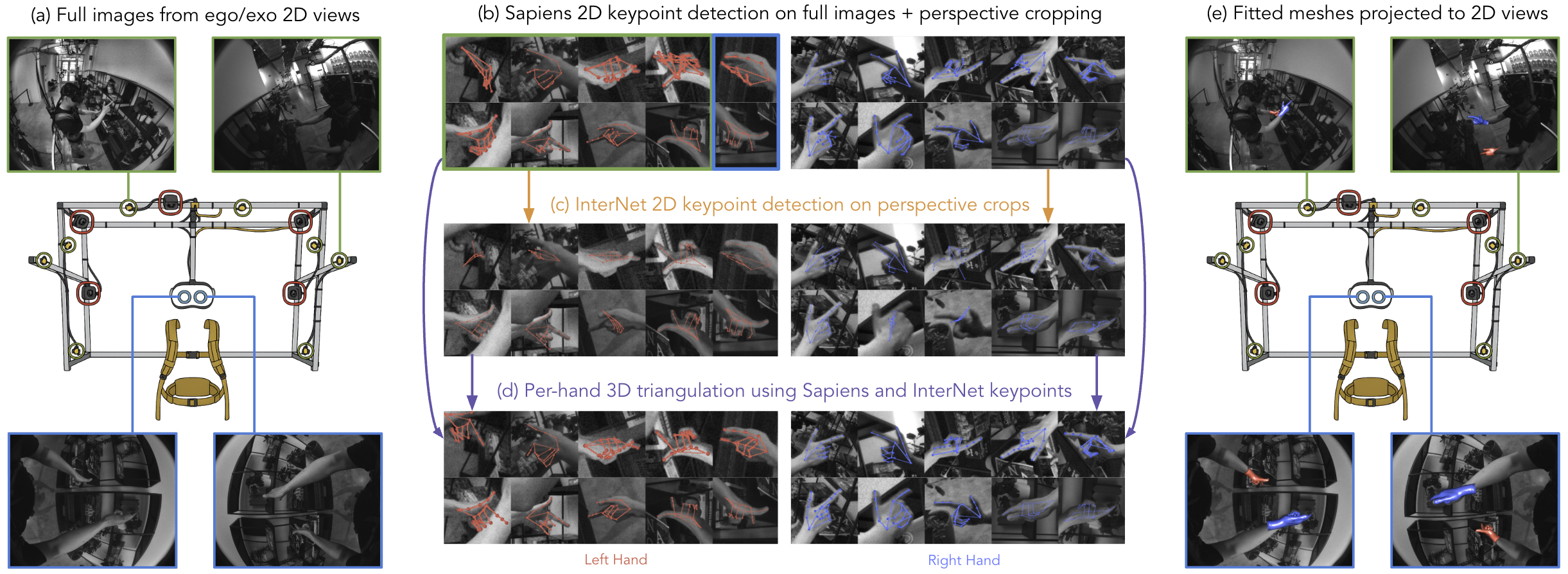}
  \vspace{-6mm}
  \caption{\textbf{Overview of our multi-stage pipeline for accurate 3D hand pose annotation.} (a) Multi-view fisheye images from our 10 ego/exo cameras. (b) Sapiens body keypoint detector predicts 2D hand keypoints per frame and localizes hands for cropping. (c) InterNet hand-specific detector on perspective crops for additional 2D hand keypoints. (d) Left/right 3D hand keypoints are triangulated across all views using robust multi-view geometry. (e) Personalized meshes are fitted to triangulated 3D keypoints and projected back to 2D views.}
  \label{fig:pipeline}
  \vspace{-3mm}
\end{figure*}

\section{Data Collection}

\subsection{Mobile Capture Rig}

We designed a portable, self-contained, backpack-style multi-camera capture rig weighing about 8 kilograms, which can be comfortably worn by the participant without significantly restricting their natural range of motion during a capture session.

Eight monochrome fisheye cameras ($1024\times1280$ resolution, 60 Hz, 152$\degree$ horizontal and 116$\degree$ vertical field of view) are rigidly mounted on the rig in a half-dome configuration to maximize visual coverage. The arrangement includes two top-down cameras on the top bar of the rig, two in the farthest corners, two on lateral protruding bars and angled inward, and two bottom-up cameras at the base of the rig (see Figure~\ref{fig:rig}). Exact camera placements and viewing angles were determined through empirical testing.

The participants also wear a Meta Quest 3 headset equipped with two identical monochrome fisheye cameras, offering egocentric views. 
As the Quest 3 is the best-selling consumer headset globally, data captured with its camera configuration closely reflects real-world usage scenarios. 
We do not fix the headset to the rig, to allow for free head movements.
To track the headset pose relative to the exocentric capture rig, we employ MoCap the same way as in HOT3D~\cite{banerjee2025hot3d}: a lightweight marker tree (rigid body) is placed atop the headset, which is tracked via five OptiTrack PrimeX 13W cameras distributed around the rig.
This is an alternative headset tracking solution to using AR tags, and generally has better precision.

All ten monochrome cameras and the five MoCap cameras are hardware-synchronized and precisely calibrated into a shared three-dimensional reference frame. 
Note that the reference frame moves with the participant, instead of being fixed to the physical world.
Video and MoCap data are streamed in real-time to a high-performance desktop workstation placed on a mobile cart. This cart easily moves alongside the participant, enabling mostly unrestricted mobility during capture.

\subsection{Automated Hand Annotation}

We present our multi-stage pipeline for accurate ground-truth annotation of hand poses that leverages both body and hand 2D keypoint detection models, followed by cross-view triangulation and personalized mesh fitting.

First, for each frame from our ten fisheye monochrome cameras (eight exocentric and two egocentric), we employ a state-of-the-art human keypoint detection model, Sapiens~\cite{khirodkar2024sapiens}, which predicts 308 body keypoints, including 42 dedicated hand keypoints (21 per hand). In addition to providing high-quality hand pose estimations, this step also serves to localize the hand in the frame. We use a version of Sapiens finetuned on large sets of proprietary human keypoint data, including data curated from publicly available sources (e.g., Shutterstock) and extensive amounts of synthetic data that can roughly approximate the real-world conditions that we encounter during our in-the-wild captures. 

Given the hand keypoint predictions from Sapiens, we next perform perspective cropping \cite{han2022umetrack} of each hand (left and right) within each camera view. 
To achieve this, for every hand, we compute the centroid of its 21 predicted keypoints and define a bounding circle centered on this centroid. 
Subsequently, we create a virtual pinhole camera aligned precisely with the physical camera's location; 
and adjust its intrinsics so that all points within the bounding circle project within a $256\times 256$ pixel perspective image captured by the virtual camera. 
This particular resolution was selected to match the input size expected by our second-stage hand-specific keypoint detection model, which we describe next.

Once perspective hand crops are obtained, we apply a dedicated 2D hand pose estimation model, InterNet~\cite{Moon_2020_ECCV_InterHand2.6M}, which was trained on a combination of proprietary datasets containing hands captured in large multi-camera domes, curated hands from public data, and synthetic data rendered to simulate complex real-world hand appearances. 
Prior to inference, we perform histogram equalization~\cite{10.5555/1076432} on each perspective-cropped hand image, mitigating challenges introduced by varying lighting.

Now, at each time step, we have two independent sets of hand keypoint predictions (Sapiens and InterNet) from each of the ten synchronized camera views, resulting in a total of 20 sets of 42 hand keypoints per frame. Both detection models provide confidence scores for each predicted keypoint, enabling a confidence-based filtering step where predictions with confidence values below a threshold of 0.3 are discarded. 
These filtered 2D detections serve as input to standard RANSAC-based triangulation, producing 42 3D keypoints for each frame.

Finally, the set of triangulated 3D hand keypoints is used to fit detailed hand meshes. 
We use a solution compatible with UmeTrack~\cite{han2022umetrack} and HOT3D~\cite{banerjee2025hot3d}, where a personalized linear blend skinning model of the hand is first derived from a high-resolution hand scan system, 
which can then be fit to the 3D keypoints via Inverse Kinematics.
This personalization ensures high-quality ground-truth meshes suitable for fine-grained tasks, such as dense contact analysis.

\subsection{Quantitative Evaluation}

To assess the accuracy of our ground-truth hand pose annotations, we conduct a controlled evaluation inside a large multi-camera capture dome equipped with 30 high-resolution cameras. This dome setup provides denser coverage and more favorable viewpoints compared to our mobile rig, and therefore serves as the ``gold standard'' to evaluate the quality of hand poses from our system. 
We synchronize and calibrate all cameras from both systems to the same coordinate frame, allowing us to directly compare our 3D hand keypoints against those obtained from the dome.


\begin{table}[h]
\centering
\begin{tabular}{lcc}
\toprule
\textbf{MPJPE (mm)} & \textbf{Median} & \textbf{P90} \\
\midrule
No interactions & 6.28 & 7.31 \\
Hand-hand interaction & 7.46 & 8.95 \\
Hand-object interaction & 7.90 & 12.53 \\
\bottomrule
\end{tabular}
\vspace{2mm}
\caption{Quantitative evaluation of our ground-truth hand annotations against a high-coverage 30-camera dome setup. We report the median and 90th percentile of per-frame mean per-joint position error (MPJPE) in millimeters.}
\label{tab:quantitative}
\end{table}

\cref{tab:quantitative} shows that across all conditions, our mobile rig achieves sub-centimeter median accuracy, closely approaching that of the dome system, despite being far more lightweight and mobile. 

Beyond dome-based validation, we also evaluate the utility of our dataset as a benchmark for 3D hand pose estimation. We build an evaluation set, which we refer to as \textbf{EgoExo-Hands}, containing around 30,000 annotated frames captured from 3 subjects in several indoor and outdoor environments.
We benchmark generalization by training the UmeTrack hand tracker~\cite{han2022umetrack} on two existing datasets---UmeTrack~\cite{han2022umetrack} and HOT3D~\cite{banerjee2025hot3d}---and testing on EgoExo-Hands. Table~\ref{tab:cross} reports the mean keypoint position error (MKPE, in mm).

\begin{table}[h]
\centering
\begin{tabular}{ccc}
\toprule
\textbf{Train set} & \textbf{Test set} & \textbf{MKPE} $\downarrow$ \\ 
\midrule
UmeTrack + HOT3D & UmeTrack & 9.48 \\ 
UmeTrack + HOT3D & HOT3D & 10.95 \\ 
\midrule
UmeTrack + HOT3D & EgoExo-Hands & 16.28 \\ 
HOT3D & EgoExo-Hands & 22.59 \\ 
UmeTrack & EgoExo-Hands & 24.78 \\ 
\bottomrule
\end{tabular}
\vspace{2mm}
\caption{Evaluation of 3D hand pose estimation. Models trained on existing datasets, UmeTrack~\cite{han2022umetrack} and HOT3D~\cite{banerjee2025hot3d}, generalize significantly worse to EgoExo-Hands, highlighting the increased difficulty of our in-the-wild data.}
\label{tab:cross}
\end{table}

As shown in \cref{tab:cross}, models trained on UmeTrack or HOT3D achieve reasonable performance when tested within their respective domains. However, when evaluated on EgoExo-Hands, errors increase substantially. 
This highlights the greater diversity and difficulty of our dataset.

The above quantitative evaluation results demonstrate that our proposed dataset fills an important gap in the existing literature: it provides a high-quality benchmark for studying robust hand pose tracking under realistic, unconstrained conditions. 






\section{Conclusion and Future Work}

We introduce a novel, mobile multi-camera capture rig to capture egocentric hand-object interactions in the wild. 
The resulting dataset, EgoExo-Hands, features accurate 3D hand pose annotations in challenging environments.

While this extended abstract focuses on accurate 3D hand pose annotation, we are in the process of collecting and building a large-scale dataset that captures the full spectrum of in-the-wild hand-object interactions. Beyond triangulated 3D hand keypoints and personalized hand meshes, this dataset will provide additional rich annotations, including: 
6DoF object pose, segmentation and contact annotations, and detailed text descriptions.

%
%

{
    \small
    \bibliographystyle{ieeenat_fullname}
    \bibliography{bibtex/bibtex}
}

\end{document}